\documentclass[runningheads]{llncs}
\usepackage{graphicx}

\usepackage{tikz}
\usepackage{comment}
\usepackage{amsmath,amssymb} 
\usepackage{color}

\usepackage{algorithm}
\usepackage{algorithmic}
\usepackage{hhline}
\usepackage{arydshln}

\usepackage{newfloat}
\usepackage{listings}
\floatstyle{ruled}
\newfloat{listing}{tb}{lst}{}
\floatname{listing}{Listing}
\usepackage{color}
\usepackage{colortbl}
\usepackage{times}
\usepackage{epsfig}
\usepackage{graphicx}
\usepackage{amsmath}
\usepackage{amssymb}
\usepackage[pagebackref=true,breaklinks=true,letterpaper=true,colorlinks,bookmarks=false,citecolor=blue,linkcolor=blue]{hyperref}

\hypersetup{
    colorlinks=true,
    urlcolor=magenta,
    citecolor=blue,
}

\makeatletter\if@twocolumn\PassOptionsToPackage{switch}{lineno}\else\fi\makeatother

\usepackage[T1]{fontenc}
\usepackage{multirow}
\usepackage{booktabs}
\usepackage{comment}
\usepackage{color}
\usepackage{subcaption}
\usepackage{textcomp}
\usepackage{siunitx}
\usepackage{tabulary}

\makeatletter
\newcommand\figref{Fig.~\ref}
\newcommand{\tabref}[1]{Table~\ref{#1}}
\let\ts@includegraphics\includegraphics

\usepackage{float}
\begin{document}
\pagestyle{headings}
\mainmatter
\def\ECCVSubNumber{4395}  

\title{RVSL: Robust Vehicle Similarity Learning in Real Hazy Scenes Based on Semi-supervised Learning} 

\titlerunning{ECCV-22 submission ID \ECCVSubNumber} 
\authorrunning{ECCV-22 submission ID \ECCVSubNumber} 
\author{Anonymous ECCV submission}
\institute{Paper ID \ECCVSubNumber}


\titlerunning{Robust Vehicle Similarity Learning in Real Hazy Scenes}
%
\author{Wei-Ting Chen\inst{1*} \and
I-Hsiang Chen\inst{2*} \and
Chih-Yuan Yeh\inst{2}\and
Hao-Hsiang Yang\inst{2}\and
Hua-En Chang\inst{2}\and
Jian-Jiun Ding\inst{2}\and
Sy-Yen Kuo\inst{2}}
\authorrunning{Chen \& Chen et al.}
%

\institute{Graduate Institute of Electronics Engineering, National Taiwan University, Taiwan \and Department of Electrical Engineering, National Taiwan University, Taiwan\\\email{\{f05943089,f09921058,f09921063\}@ntu.edu.tw,\\islike8399@gmail.com,\{r10921a35,jjding,sykuo\}@ntu.edu.tw}
\\ \url{https://github.com/Cihsaing/rvsl-robust-vehicle-similarity-learning--ECCV22}}

\maketitle

\begin{abstract}
Recently, vehicle similarity learning, also called re-identification (ReID), has attracted significant attention in computer vision. Several algorithms have been developed and obtained considerable success. However, most existing methods have unpleasant performance in the hazy scenario due to poor visibility. Though some strategies are possible to resolve this problem, they still have room to be improved due to the limited performance in real-world scenarios and the lack of real-world clear ground truth. Thus, to resolve this problem, inspired by CycleGAN, we construct a training paradigm called \textbf{RVSL} which integrates ReID and domain transformation techniques. The network is trained on semi-supervised fashion and does not require to employ the ID labels and the corresponding clear ground truths to learn hazy vehicle ReID mission in the real-world haze scenes. To further constrain the unsupervised learning process effectively, several losses are developed. Experimental results on synthetic and real-world datasets indicate that the proposed method can achieve state-of-the-art performance on hazy vehicle ReID problems. It is worth mentioning that although the proposed method is trained without real-world label information, it can achieve competitive performance compared to existing supervised methods trained on complete label information.

\end{abstract}

\newcommand\blfootnote[1]{%
\begingroup
\renewcommand\thefootnote{}\footnote{#1}%
\addtocounter{footnote}{-1}%
\endgroup
}
\blfootnote{*Indicates equal contribution.}

\section{Introduction}
Vehicle similarity learning, also called vehicle re-identification (ReID), is a crucial technique for intelligent surveillance systems in a smart city. It is to track the vehicles with the same identity within a set of images captured by multiple cameras and various viewpoints. With the development of the deep convolutional neural network (DCNN), several approaches for vehicle ReID~\cite{he2019part,he2021transreid,meng2020parsing,zhao2021heterogeneous} have been proposed and achieved impressive performance. Similar to other high-level vision applications such as object detection and semantic segmentation~\cite{chen2021all}, although existing methods can handle vehicle ReID effectively on normal images, they have limited performance under inclement weather, especially in hazy scenario. Haze is a common and inevitable weather phenomenon that leads to poor visual appearances and causes the loss of discriminative information by deteriorating the contents of images for vehicle ReID. Thus, this field still has room for improvement.

Inspired by previous dehazing tasks~\cite{cai2016dehazenet}, we can apply an atmospheric scattering model~\cite{koschmieder1924theorie} to synthesize haze images and then train the vehicle ReID models based on the rendered images and the corresponding ID labels. Though this strategy can achieve decent performance on synthetic images, they have limited performance on real haze images due to the domain gap between synthetic and real-world images~\cite{chen2022sjdl}. While this issue can be resolved by adopting real haze images in the training stage, collecting the real haze data and labeling the correct ground truths are difficult and troublesome.

Another possible baseline strategy is to adopt the existing dehazing approaches~\cite{chen2019pms,zhang2018densely} or comprehensive image restoration method~\cite{zamir2021multi} as the pre-processing technique and then apply the ReID. Although the above strategies are shown to achieve promising results in haze removal, there is no guarantee that the selected pre-processing techniques would be able to improve ReID, since these two tasks are performed separately and existing dehazing methods are not designed for the purpose of ReID but for human perception. Moreover, most existing dehazing methods require pair data to train the model, but it is infeasible to attain the ground truths of haze images in real-world scenes. Though we can adopt synthetic data to train the network, the domain gap problem may still exist, which may generate undesired dehazed results in real haze scenes and further limit the performance of ReID.

\begin{figure*}[t!]
\centering \includegraphics[width=1.0\textwidth]{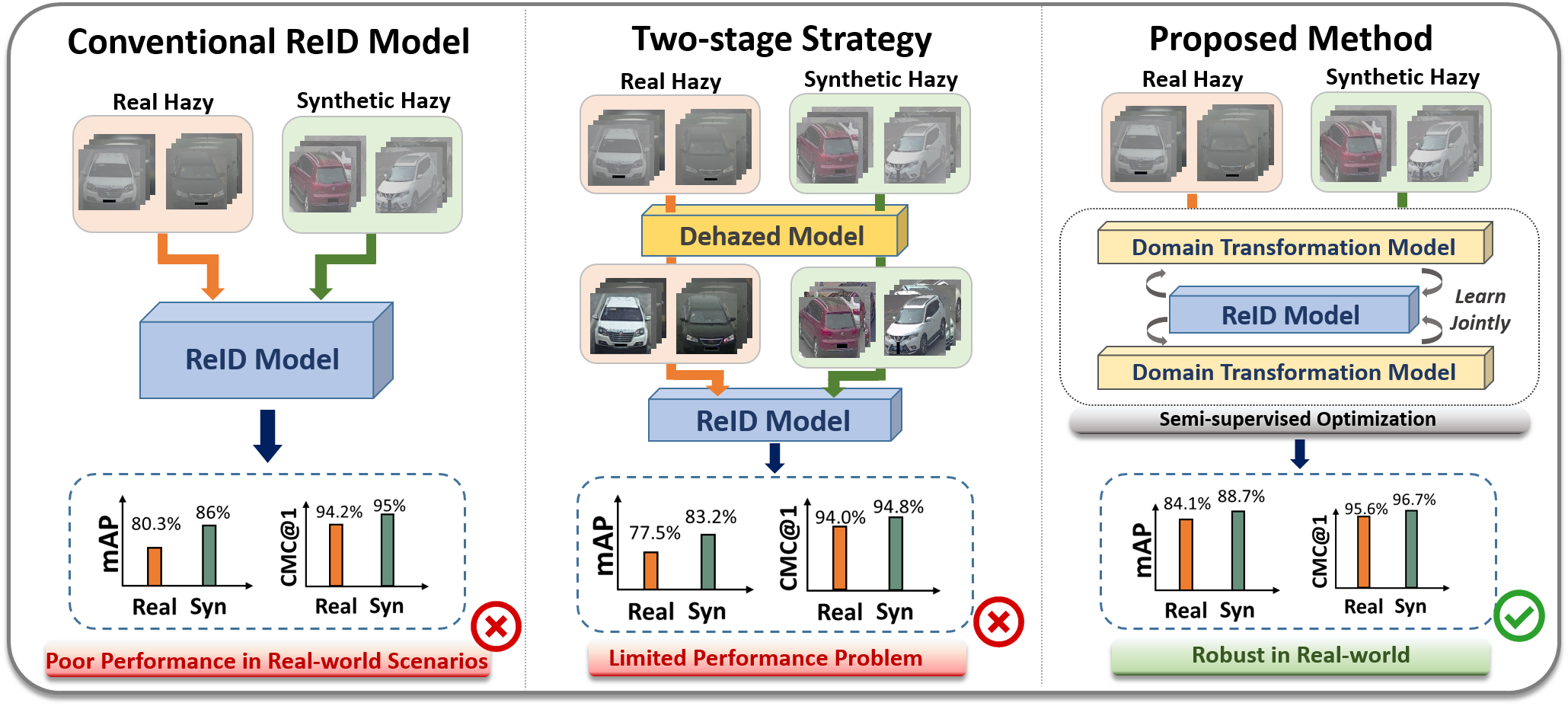}{}
\makeatother 
\caption{\textbf{Illustration of different strategies to solve hazy vehicle ReID problem.} One can see that our method outperforms other existing methods in terms of the mean average precision (mAP) and CMC@1. Moreover, other strategies may have limited performance problem in real-world scenarios. We adopt CAL~\cite{rao2021counterfactual} and MPR-Net~\cite{zamir2021multi} as the ReID model and the dehazed model, respectively.}
\label{fig:teasor_fig}
\end{figure*}

By the above analysis, there are two reasons that hinder the development of ReID in real haze scenes: (i) the scarcity of labels of real-world data and (ii) the lack of appropriate guidance for real haze. In this paper, to mitigate these problems, we construct a novel training architecture based on the deep convolutional neural network (DCNN). Inspired by CycleGAN~\cite{isola2017image} which can transform images between any two domains, we introduce the domain transfer technique in the proposed network and combine it with the vehicle ReID. Specifically, the proposed method is trained on a semi-supervised paradigm in an end-to-end fashion and there are two parts in the training process: supervised training for synthetic data and unsupervised training for real-world data. For the former part, the network can learn the knowledge of transformation between two domains and extract more discriminative features for vehicle ReID in fully supervision by paired data (i.e., synthetic hazy images and the corresponding clear ground truths). For the latter parts, we only leverage two sets of unpaired data (i.e., real hazy images and clear images) to strengthen the robustness of the domain transformation and the ReID in real-world scenes in an unsupervised manner.

The idea of our method is that, the domain transformation network transfers the input image (i.e., hazy or clear images) between two domains with the same background information and the ReID network extracts the latent features for classifying from two images (i.e., input and transferred image). Inspired by the cycle consistency~\cite{isola2017image,yan2020optical}, two extracted embedding features should be identical since they are from the same vehicles. Thus, we can calculate the consistency of between two extracted features for optimizing the network. 

Based on the semi-supervised training scheme, the utilization of the synthetic data can guide the unsupervised stage and prevent the network from unstable performance~\cite{isola2017image}. On the other hand, the use of real-world data can improve the generalization ability of our model to real data and further mitigate the domain gap problem when the synthetic data is applied in the training process~\cite{li2019semi}. Moreover, using the domain transformation network can assist the ReID network to learn more discriminative features for the ReID under real-world haze scenarios. By our design, the proposed method can perform vehicle ReID in hazy scenarios effectively without additional annotations in real hazy data which are usually hard to be obtained. Furthermore, our proposed training scheme can be also applied in the case that we have annotations of vehicle ID and achieve better performance.

The contribution of this paper is summarized as follows.
\begin{itemize}
\itemsep0em 
\item {A novel training paradigm based on semi-supervised learning and domain transformation is proposed to learn hazy vehicle ReID without the labels or clear ground truths of real-world data. We term it \textbf{R}obust \textbf{V}ehicle \textbf{S}imilarity \textbf{L}earning (\textbf{RVSL}). As depicted in \figref{fig:teasor_fig}, by combining domain transferring technique with the ReID network, the proposed method can achieve decent performance to learn discriminative features under real haze scenes without using ID label. Surprisingly, the proposed method achieves competitive performance compared with other existing methods trained with complete ID information.}
\item {To constrain the unsupervised stage in the training process, we developed several loss functions such as embedding consistency loss, colinear relation constraint, and monotonously increasing dark channel loss to improve the performance. These loss functions enable the network to learn both domain transformation and ReID in an unsupervised way effectively. Experimental results prove the effectiveness of these loss functions.}
\end{itemize}

\section{Related Works}
\noindent
\textbf{Vehicle Re-identification.} With the great effort of data collection and annotation, several large-scale benchmarks for vehicle ReID such as VehicleID~\cite{liu2016deep}, VeRi-776~\cite{liu2017provid}, VERI-Wild~\cite{lou2019veri}, and Vehicle-1M~\cite{guo2018learning}) are proposed. Based on these well-developed benchmarks, several approaches~\cite{hermans2017defense,gao2020vehicle,he2020multi,li2021self,rao2021counterfactual,wang2017orientation} have been developed and most of them rely on DCNN. We can divide them into the following categories.
\noindent \smallskip\\
1) \textit{Meta-information-based methods} which integrate meta-information for feature learning. For example, Zheng \textit{et al.}~\cite{zheng2019attributes} leveraged the additional information such as the camera view, and the vehicle type and color to guide the network. Shen \textit{et al.}~\cite{shen2017learning} integrated the visual-spatio-temporal path proposals and spatial-temporal relations to a Siamese-CNN+Path-LSTM network.
Rao \textit{et al.}~\cite{rao2021counterfactual} proposed the attention mechanism with counterfactual causality which enables the network to learn more useful attention for fine-grained features for ReID.
\noindent \smallskip\\
2) \textit{Local information-based methods}: Meng \textit{et al.}~\cite{meng2020parsing} adopted the common region information extracted by a vehicle part parser to improve the mutual representation information between different viewpoints. Khorramshahi \textit{ et al.}~\cite{khorramshahi2020devil} applied Variational Auto-Encoder (VAE) to find crucial detailed information which can be regarded as the pseudo-attention map for highlighting discriminative regions. He \textit{et al.}~\cite{he2019part} combined local and non-local information based on a part-regularized mechanism. These strategies can preserve the variance from near-duplicate vehicles to improve the performance of vehicle ReID. Zhao \textit{et al.}~\cite{zhao2021heterogeneous} applied Cross-camera Generalization Measure technique and integrated region-specific features and cross-level features together to improve the performance of ReID. 
\noindent \smallskip\\
3) \textit{Generative Adversarial Network (GAN) based methods}: Zhou \textit{et al.}~\cite{zhou2018aware} applied the conditional multi-view generative network to extract global feature representation from various viewpoints and then adopted adversarial learning to facilitate feature generation. Lou \textit{et al.}~\cite{lou2019veri} designed the FDA-Net to generate hard examples in the feature space based on the GAN to improve the robustness of ReID. Yao \textit{et al.}~\cite{yao2020simulating} proposed to adopt a 3D graphic engine to reduce the content gap between the existing datasets to suppress the domain gap problem. 
\noindent \smallskip\\
4) \textit{Vision Transformer (ViT) based Methods}: He \textit{et al.}~\cite{he2021transreid} leveraged the ViT to encode input images as a vector for embedding representation. To further improve representation learning, the jigsaw patch module and side information were adopted in the training scheme. 

Though the above methods can achieve decent vehicle ReID performance on clear images, they are still limited in real-world hazy image scenarios.

\noindent \smallskip\\
\textbf{Single Image Haze Removal.} Based on Koschmieder's model \cite{koschmieder1924theorie}, the formation of haze can be modeled by:
\begin{equation}
I\left(x\right)=J\left(x\right)t\left(x\right)+A\left(1-t\left(x\right)\right),
\label{eq:fog model}
\end{equation}
where $I(x)$ is the hazy image, $J(x)$ is the haze-free image and $A$ is the global atmospheric light. $t\left(x\right)=e^{-\beta d\left(x\right)}$ is the medium transmission map where $\beta$ is the scattering coefficient and $d(x)$ is the depth from the camera to the object. There are numerous haze removal methods proposed in past decades. They can be classified into prior-based and deep learning-based methods. The former class is to explore the prior knowledge between hazy and haze-free images. For example, He \textit{et al.}~\cite{he2010single} proposed the dark channel prior, Zhu \textit{et al.}~\cite{zhu2015fast} developed the color attenuation prior, and Berman \textit{et al.} proposed the haze-line~\cite{berman2016non} to estimate the dehazed results. The other class is to apply the DCNN. For instance, Qu \textit{et al.}~\cite{Qu_2019_CVPR} proposed multi-resolution generators and discriminators for dehazing in a coarse-to-fine way. Dong \textit{et al.}~\cite{dong2020multi} used the strengthen-operate-subtract boosting strategy to improve the dehazing network. Wu \textit{et al.}~\cite{wu2021contrastive} proposed an auto-encoder-like framework with additive mixup operation and a dynamic feature enhancement module to improve the quality of extracted features for dehazing. Zamir \textit{et al.}~\cite{zamir2021multi} proposed a multi-stage architecture that can encode a diverse set of features simultaneously to restore accurate outputs. Chen \textit{et al.}~\cite{chen2022learning} proposed a unified architecture which can learn multiple adverse weather based on a single architecture.

\begin{figure*}[t!]
\centering \includegraphics[width=1.0\textwidth]{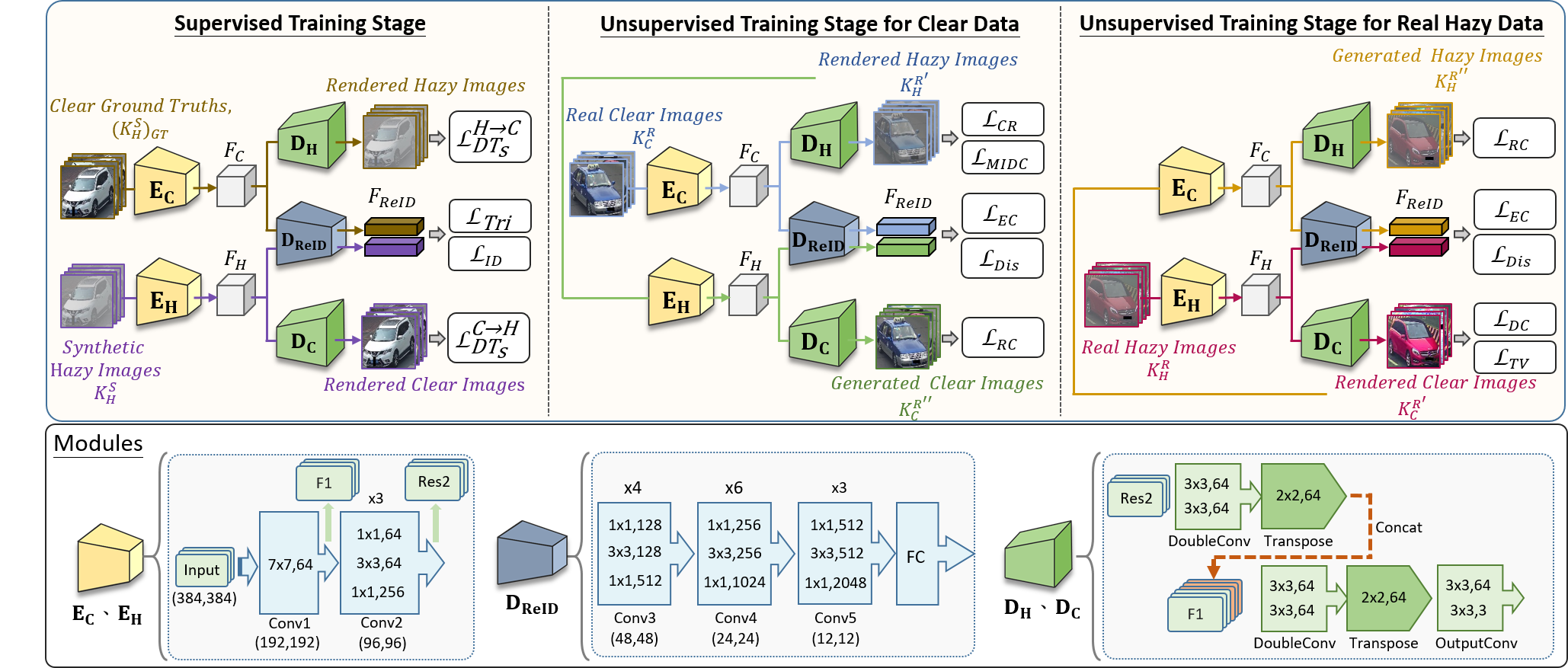}{}
\makeatother 
\caption{\textbf{The architecture of the proposed semi-supervised hazy vehicle ReID network.} Our method consists of supervised and unsupervised training stages for synthetic and real-world data.}
\label{fig:architecture}
\end{figure*}

\section{Proposed Method}
\subsection{Overview of the Proposed Method}
As shown in \figref{fig:architecture}, there are five modules in the proposed network. That is, two encoders for hazy and clear scenes ($\mathbf{E_{H}}$ and $\mathbf{E_{C}}$), and three decoders for hazy, clear, and ReID ($\mathbf{D_{H}}$, $\mathbf{D_{C}}$, and $\mathbf{D_{ReID}}$), respectively. These modules can be combined to two sub-networks called domain transformation network and re-identification network. The features extracted by $\mathbf{E_{H}}$ and $\mathbf{E_{C}}$ are termed $F_{H}$ and $F_{C}$, respectively.

As mentioned in section 1, due to the lack of real haze data, in this paper, inspired by semi-supervision~\cite{li2019semi}, we apply both synthetic data and real data simultaneously in the learning process. At the supervised training stage, the application of the synthetic data can learn the transformation from various domains stably. At the unsupervised stage, we first take the real clear images and then take real hazy images as inputs, respectively. This operation enables our network to learn real-world information of hazy scenes, clear scenes, and ReID through the domain transformation network and ReID network simultaneously. We illustrate the details in the following subsections.

\subsection{Domain Transformation Network.}
The goal of the domain transformation network (DT-Net) is to help the ReID network to learn haze-invariant features via transforming the domain of the input data. The detailed illustration of this network is as follows.
\\\\
\noindent\textbf{Architecture.} The DT-Net consists of two encoders ($\mathbf{E_{H}}$ and $\mathbf{E_{C}}$) and two decoders ($\mathbf{D_{H}}$ and $\mathbf{D_{C}}$). Given a hazy input, the decoder $\mathbf{D_{C}}$ generates the corresponding clear image based on the features $F_{H}$ extracted by encoder $\mathbf{E_{H}}$. On the other hand, the decoder $\mathbf{D_{H}}$ takes the features $F_{C}$ extracted from the encoder $\mathbf{E_{C}}$ to produce the hazy image. The features (i.e., $F_{C}$ and $F_{H}$) extracted by two encoders pass through a double convolution block and a deconvolution block for dimension matching. Then, the upsampled features are concatenated with the features extracted by the first convolution blocks in the encoders to improve the feature diversity. This operation is based on the fact that the features in the shallow layer of the network contain more fruitful spatial and contextual information which can benefit domain transformation~\cite{chen2021Desmoke,hui2020image} while the deeper layers usually consist of more high-level vision. The concatenated features are passed through a double convolution block and a deconvolution block to reconstruct the final domain transformation results. The quality of domain transformation is crucial for the ReID network since it may affect the feature extraction of the input image. Thus, for the synthetic data, we adopt the supervised loss $\mathcal{L}_{DTs}$ to optimize the networks. For real data, we adopt unsupervised losses $\mathcal{L}_{DTu}$.
\\\\
\noindent\textbf{Supervised Training Stage.} At this stage, we can train the network in a fully supervised way since the corresponding clear ground truths and ID labels are available. First, we adopt synthetic image pairs to train the DT-Net (i.e., $\mathbf{E_{H}}$, $\mathbf{E_{C}}$, $\mathbf{D_{H}}$, and $\mathbf{D_{C}}$). Specifically, the synthetic haze image $K_{H}^{S}$ and the corresponding clear ground truth $(K_{H}^{S})_{GT}$ are fed into the domain transformation network to calculate the domain transformation loss $\mathcal{L}_{DT_{s}}$ for synthetic data. This operation aims to constrain the distance between the predicted results (i.e., the rendered hazy images and the rendered clear images) and the corresponding ground truths. The domain transformation loss $\mathcal{L}_{DTs}$ can be formulated as follows.
\begin{equation}
    \resizebox{0.55\hsize}{!}{
    $\begin{split}
        \mathcal{L}_{DT_{s}}^{H\rightarrow C}=\frac{1}{M}\sum\limits_{i=1}^{M}\Vert \mathbf{D_{C}}[\mathbf{E_{H}}[K_{H}^{S}(i)]]-(K_{H}^{S})_{GT}(i)\Vert_{1}
    \end{split}$}
\end{equation}
  \begin{equation}
  \resizebox{0.55\hsize}{!}{
  $\begin{split}
    \mathcal{L}_{DT_{s}}^{C\rightarrow H}=\frac{1}{M}\sum\limits_{i=1}^{M}\Vert \mathbf{D_{H}}[\mathbf{E_{C}}[(K_{H}^{S})_{GT}(i)]]-K_{H}^{S}(i)\Vert_{1}
    \end{split}$}
  \end{equation}  
where $\Vert\cdot\Vert_{1}$ presents the $L_{1}$ norm and $M$ indicates the number of images. $\mathcal{L}_{DT_{s}}^{C\rightarrow H}$ and $\mathcal{L}_{DT_{s}}^{H\rightarrow C}$ denote the domain transformation loss for 'clear to haze' and 'haze to clear', respectively. The $\mathcal{L}_{DT_{s}}$ loss is the summation of $\mathcal{L}_{DT_{s}}^{C\rightarrow H}$ and $\mathcal{L}_{DT_{s}}^{H\rightarrow C}$.
\\\\
\noindent\textbf{Unsupervised Training Stage for Clear Data.} At this stage, to train the DT-Net without the hazy image ground truths, first, we adopted the cycle-consistency mechanism. The input clear image $K_{C}^{R}$ is fed into the DT-Net to render the hazy image $K_{H}^{R'}$, where $K_{H}^{R'}=\mathbf{D_{H}}(\mathbf{E_{C}}(K_{C}^{R}))$. Then, we further take the rendered image to the DT-Net to generate the rendered clear image $K_{C}^{R''}$, where $K_{C}^{R''}=\mathbf{D_{C}}(\mathbf{E_{H}}(K_{H}^{R'}))$.
In the same time, several loss functions are adopted to optimize the network. The loss at this stage $\mathcal{L}_{DT_{uc}}$ consists of the rendering consistency loss ($\mathcal{L}_{RC}$), the monotonously increasing dark channel loss ($\mathcal{L}_{MIDC}$), the colinear relation constraint ($\mathcal{L}_{CR}$), and the discriminative loss ($\mathcal{L}_{Dis}$). We illustrate each of them as follows.
\\\\
(i) \textit{Rendering Consistency Loss.} This loss is to constrain the learning process of the domain transformation network (i.e., $\mathbf{E_{H}}$, $\mathbf{E_{C}}$, $\mathbf{D_{H}}$, and $\mathbf{E_{C}}$). We adopt the pixel-wise difference between the clear input image $K_{C}^{R}$ and the rendered clear image $K_{C}^{R''}$ to ensure that the domain transformation process can be conducted in two different domains robustly. This loss is formulated as follows.
\begin{equation}
    \centering
    \resizebox{0.38\hsize}{!}{
$\begin{split}
    \mathcal{L}_{RC} = \frac{1}{M} \sum_{i=1}^{M} \vert\vert K_{C}^{R}(i)-K_{C}^{R''}(i)\vert\vert_{1}
    \label{eq:RC_loss}
    \end{split}$}
\end{equation}
\\
\smallskip
(ii) \textit{Monotonously Increasing Dark Channel Loss.} To further improve the image quality of rendered haze images, inspired by dark channel prior (DCP)~\cite{he2010single}, we propose the monotonously increasing dark channel loss $\mathcal{L}_{MIDC}$. The DCP demonstrates that for most natural clear images, the dark channel values may be close to zero. Specifically, it can be defined as:
\begin{equation}
    \centering
    \resizebox{0.45\hsize}{!}{
$\begin{split}
    \textit{DC}(\textit{J})\left(x\right)=\underset{y\in\Omega\left(x\right)}{min}\left(\underset{c\in\left\{r,g,b\right\}}{min}J^{c}\left(y\right)\right)\approxeq0, 
    \label{eq:hl_loss}
    \end{split}$
    }
\end{equation}
where \textit{DC}($\cdot$) is the operation of the dark channel, $J^{c}(y)$ is the intensity in the color channel \textit{c}, and \textit{\ensuremath{\Omega }}(\textit{x}) is a local patch with a fixed size centered at \textit{x}. With this prior, it can be further extended that, given an image deteriorated by haze, its dark channel value may be higher than that of the original clear image (i.e., $DC(I)(x)\geq DC(J)(x)$). Based on this idea, we proposed $\mathcal{L}_{MIDC}$ which is determined as:
\begin{equation}
    \centering
    \resizebox{0.6\hsize}{!}{
$\begin{split}
    \mathcal{L}_{MIDC} = \frac{1}{M} \sum_{i=1}^{M} {DM}(i) \vert\vert DC(K_{H}^{R'}(i))-DC(K_{C}^{R})(i)\vert\vert_{1}
    \label{eq:RC_loss}
    \end{split}$}
\end{equation}
where $DM(i)$ is a binary map that identifies the region where the dark channel values of the clear image $K_{C}^{R}$ are higher than that of its rendered haze result $K_{H}^{R'}$ (i.e., $DC(K_{H}^{R'})(x) < DC(K_{C}^{R})(x)$). With it, we can prevent the rendered pixels from irrational results and further improve the robustness of domain transformation.
\smallskip\\\\
(iii) \textit{Colinear Relation Constraint.} Due to the inaccessibility of the ground truth of real-world hazy images, the domain transformation process may generate undesired fake image contents without appropriate training. Although we can leverage the synthetic data to guide the network at the training stage of the synthetic data, it is not applicable at the training stage of the real-world data. To further strengthen the robustness of transformation (i.e., $\mathbf{E_{C}}$ and $\mathbf{D_{H}}$), inspired by the haze-line prior~\cite{berman2016non,yan2020optical}, we develop colinear relation constraint $\mathcal{L}_{CR}$.

Based on the physical model of haze illustrated in \eqref{eq:fog model}, Berman \textit{et al.}~\cite{berman2016non} observe that the clear image $J$, the hazy image $I$, and the atmospheric light $A$ are colinear in the RGB space (i.e., $I(x)-A=t(x)(J(x)-A)$). We can adopt this relation to constrain the training process of the network for real-world scenarios. We define the colinear relation constraint as follows.
\begin{equation}
    \centering
    \resizebox{0.6\hsize}{!}{
$\begin{split}
    \mathcal{L}_{CR} = \frac{1}{M} \sum_{i=1}^{M} \left[1-\phi(K_{C}^{R}(i)-A(i),K_{H}^{R'}(i)-A(i))\right], 
    \label{eq:hl_loss}
    \end{split}$}
\end{equation}
where $A(i)$ is the atmospheric light estimated by the rendered hazy image $K_{H}^{R'}(i)$ and $\phi(\cdot)$ means the cosine similarity. Different from~\cite{yan2020optical}, we adopt the atmospheric light estimation method in~\cite{he2010single} in this loss. With this loss, the consistency of structure and color can be further constrained.
\\\\
\smallskip
(iv) \textit{Discriminative Loss.}
To further constrain unsupervised domain transformation, we adopt the discriminative loss~\cite{goodfellow2014generative} in the training process to distinguish whether the rendered hazy image $K_{H}^{R'}$ is real or fake. In our method, we adopt the saturating discriminative loss~\cite{gui2021review}.
\\\\
\noindent\textbf{Unsupervised Training Stage for Real Hazy Data.} At this stage, the real hazy images are adopted to optimize the network without clean ground truths and ID labels. Like the previous stage, the hazy images are fed into the DT-Net (i.e., $\mathbf{E_{H}}$ and $\mathbf{D_{C}}$) to generate the clear images $K_{C}^{R'}$. Subsequently, the rendered clear images are fed into the DT-Net to obtain the rendered hazy images $K_{H}^{R''}$. To optimize our framework, apart from the monotonously increasing dark channel loss ($\mathcal{L}_{MIDC}$) and colinear relation constraint ($\mathcal{L}_{CR}$), the rest of losses in $\mathcal{L}_{DT_{uc}}$ are adopted. Moreover, to improve the predicted clear images by the DT-Net, we introduce two losses: the dark channel loss $\mathcal{L}_{DC}$ to curb the residual haze and total variation loss $\mathcal{L}_{TV}$ to prevent the noise generation. They can be formulated as follows.
\begin{equation}
  \resizebox{0.35\hsize}{!}{
  $\begin{split}
\mathcal{L}_{DC}=\frac{1}{M}\sum\limits_{i=1}^{M}\Vert DC(K_{C}^{R'}(i))\Vert_{1},
    \end{split}$}
  \end{equation}
\begin{equation}
  \resizebox{0.53\hsize}{!}{
  $\begin{split}
\mathcal{L}_{TV}=\frac{1}{M}\sum\limits_{i=1}^{M}\Vert \bigtriangledown_{x}K_{C}^{R'}(i)\Vert_{1}+\Vert\bigtriangledown_{y}K_{C}^{R'}(i)\Vert_{1},
    \end{split}$}
  \end{equation}
where $\bigtriangledown_{x}$ and $\bigtriangledown_{y}$ denote the gradient operations along the horizontal and vertical directions, respectively.

\subsection{Re-identification Network}
The re-identification network (ReID-Net) aims to extract discriminative features to search the images with the same identification in the gallery. The details of architecture and training are illustrated as follows.
\\\\
\noindent\textbf{Architecture.} The ReID-Net consists two encoders (i.e., $\mathbf{E_{H}}$ and $\mathbf{E_{C}}$) and one decoder (i.e., $\mathbf{D_{ReID}}$). It adopts ResNet-50~\cite{he2016deep} as the backbone, where we apply the first two convolution blocks as the architecture of two encoders. As shown in~\figref{fig:architecture}, the extracted features $F_{H}$ or $F_{C}$ are fed to the decoder $\mathbf{D_{ReID}}$ to generate the ReID results where the decoder consists of the rest of convolution blocks in ResNet-50 and extracted features are down-scaled by global average pooling (GAP) and batch normalization (BN) to generate 2048-d embedding features $F_{ReID}$. Last, we adopt the fully connected layer (FC layer) to match the number of identities for the classification. For the supervised learning of synthetic data, since we have the corresponding ID label, we adopt the triplet loss $\mathcal{L}_{Tri}$ and the ID loss $\mathcal{L}_{ID}$. For the unsupervised learning stage, due to lack of the ID label, the embedding consistency loss $\mathcal{L}_{EC}$ is adopted to constrain the network. This architecture enables our two encoders to learn domain adaptive features because the features extracted by two encoders working on different domains are fed into the same decoder.
\\\\
\noindent\textbf{Supervised Training Stage.} At this stage, we train the re-identification network (i.e., $\mathbf{E_{C}}$, $\mathbf{E_{H}}$, and $\mathbf{D_{ReID}}$) by adopting the triplet loss~\cite{hermans2017defense} $\mathcal{L}_{Tri}$ and the ID loss $\mathcal{L}_{ID}$ which can be defined as follows.
  \begin{equation}
  \resizebox{0.7\hsize}{!}{
  $\begin{split}
    \mathcal{L}_{Tri}=\frac{1}{M} \sum_{i=1}^{M}\sum_{k}\left[\max_{z_{p}\in\mathcal{P}(z_{i}^{k})}D(z_{i}^{k},z_{p})-\min_{z_{n}\in\mathcal{N}(z_{i}^{k})}D(z_{i}^{k},z_{n})+\delta \right]_{+}
    \end{split}$}
    \label{eq:triploss}
  \end{equation}

\begin{equation}
    \resizebox{0.4\hsize}{!}{
    $\begin{split}
        \mathcal{L}_{ID} = - \frac{1}{M} \sum_{i=1}^{M}\sum_{k} \log \frac{\exp(\sigma_{i}^{y_{i}^{k}})}{\sum_{j = 1}^{C} \exp(\sigma_{i}^{j})}
    \end{split}$}
  \label{eq:idloss}
\end{equation}
where $k\in\{(K_{H}^{S})_{GT},K_{H}^{S}\}$. $\mathcal{P}(z_{i}^{k})$ and $\mathcal{N}(z_{i}^{k})$ denote the positive and negative sample sets, respectively. $z_{i}^{k}$ represents the extracted embedding features from the $i^{th}$ input sample (i.e., ($(K_{H}^{S})_{GT}(i)$ or $K_{H}^{S}(i)$)). $\delta$ is the margin of the triplet loss, $D(\cdot,\cdot)$ is the Euclidean distance, and $[\cdot]_{+}$ equals to $max(\cdot,0)$. For $\mathcal{L}_{ID}$, $\sigma_{i}^{j}$ is the output of the FC layer with the class $j$ based on $i^{th}$ input image. $C$ presents the total number of the class, and $y_{i}$ donates the ground truth class. The ReID loss $\mathcal{L}_{ReID_{s}}$ at this stage is the combination of $\mathcal{L}_{ID}$ and $\mathcal{L}_{Tri}$.
\\\\
\noindent\textbf{Unsupervised Training Stage.} At this stage, we feed both real clear data and real hazy data separately. Due to the lack of labels about ID information, to train the ReID network (i.e., $\mathbf{E_{C}}$, $\mathbf{E_{H}}$, and $\mathbf{D_{ReID}}$) with real clear inputs, we develop \textit{embedding consistency loss} ($\mathcal{L}_{EC}$) to calculate the distance of two embedding features extracted from the input clear image and the rendered haze image. Initially, given a clear image $K_{C}^{R}$, it is fed into the DT-Net to render the hazy image $K_{H}^{R'}$ and ReID-Net to extract embedding feature $(F_{ReID})_{C}^{R}$. Then, we further take the rendered image to the ReID-Net to produce the embedding feature $(F_{ReID})_{H}^{R'}$. We can calculate the loss between $(F_{ReID})_{H}^{R'}$ and $(F_{ReID})_{C}^{R}$ because they are the same vehicle. By using this loss, the haze-invariant features can be learned by the ReID-Net effectively. The mathematical expression of this loss is defined as follows.
\begin{equation}
    \centering
    \resizebox{0.5\hsize}{!}{
$\begin{split}
    \mathcal{L}_{EC} = \frac{1}{M} \sum_{i=1}^{M} \vert\vert [F_{ReID}(i)]_{C}^{R}-[F_{ReID}(i)]_{H}^{R'}\vert\vert_{1}
    \label{eq:EC_loss}
    \end{split}$}
\end{equation}
By contrast, when the input data is a real hazy image (i.e., $K_{H}^{R}$), the same mechanism is adopted.

\section{Experiments}
\subsection{Dataset and Evaluation Protocols}
\begin{figure}[t!]
\centering \includegraphics[width=0.6\textwidth]{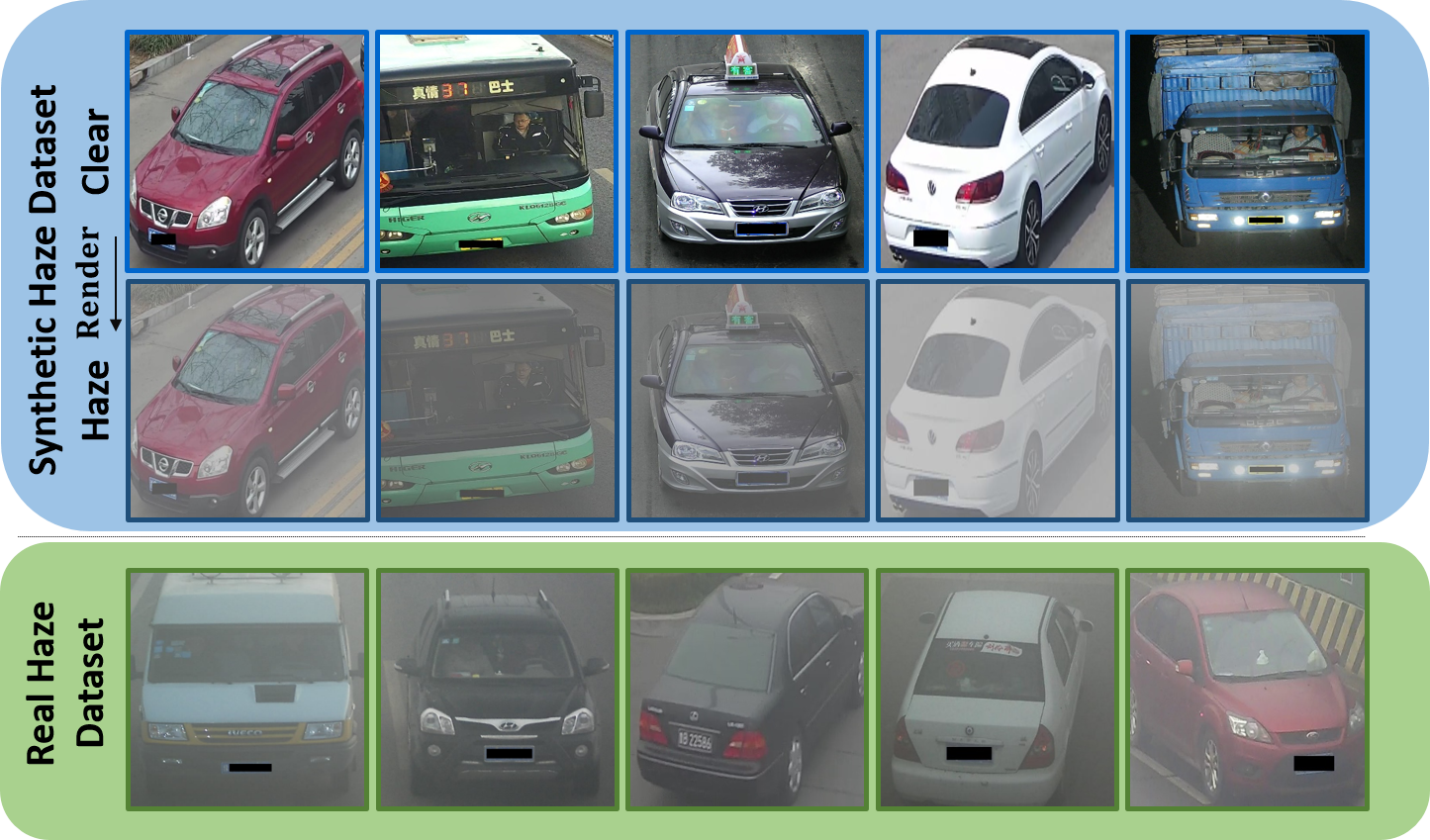}{}
\makeatother 
\caption{\textbf{Examples of the images in the synthetic dataset and the real-world datasets for vehicle ReID.}}
\label{fig:dataset-ex}
\end{figure}

\noindent
\textbf{Dataset Preparation.} The proposed semi-supervised scheme is trained by both synthetic and real-world haze data. We select haze-free images from Vehicle-1M and VERI-Wild datasets. Subsequently, we apply the haze synthesis procedure proposed in~\cite{li2018benchmarking} to synthesize these images. First, we adopt the method in~\cite{liu2015learning} to estimate the depth map $d$. Then, we render the haze on these clear images by \eqref{eq:fog model} with the predicted depth maps and set $\beta\in[0.4, 1.6]$ and $A\in[0.5, 1]$. Uniquely, each clear data generates a hazy image and all rendered images are divided into the training and the testing sets, respectively. For the real haze data, we survey all existing datasets and find that only Vehicle-1M and VERI-Wild datasets contain the cases in the hazy weather. Thus, we carefully select the vehicle images under hazy scenarios from two datasets. The selected images are split to the training and the testing sets. The details and examples of two types of data are presented in \tabref{tab:syndataset}, \tabref{tab:realdataset} and \figref{fig:dataset-ex}, respectively.
\noindent \smallskip\\
\textbf{Evaluation Protocols.} Followed by the protocols of the evaluation proposed in~\cite{lou2019veri,guo2018learning}, we randomly select one hazy image for each vehicle and put it into the probe set. The remained images form the gallery set. We adopt the cumulative matching characteristic (CMC) curve and mean average precision (mAP) to evaluate the performance.

\begin{figure}[t!]
  \begin{minipage}[t]{1.0\textwidth}
  \begin{minipage}[t]{0.5\textwidth}
    \captionof{table}{\textbf{Detail of the synthetic dataset.} (IDs/Images)}
        \centering
        \label{tab:syndataset}

        \scalebox{0.7}{
        \begin{tabular}{cccc} 
\toprule
\textbf{Set} & \textbf{Train} & \textbf{Probe} & \textbf{Gallery} \\ 
\hline\hline
\textbf{VERI-Wild} & 1167/19532 & 389/389 & 389/6125 \\
\textbf{Vehicle-1M} & 1833/23026 & 611/611 & 611/7093 \\
\textbf{Total} & 3000/42558 & 1000/1000 & 1000/13218 \\
\bottomrule
\end{tabular}
}
\end{minipage}
\hspace{0.3em}
 \begin{minipage}[t]{0.5\textwidth}
      \captionof{table}{\textbf{Detail of the real-world dataset.} (IDs/Images)}
    \centering
    \label{tab:realdataset}

\scalebox{0.7}{
\begin{tabular}{cccc} 
\toprule
\textbf{Set} & \textbf{Train} & \textbf{Probe} & \textbf{Gallery} \\ 
\hline\hline
\textbf{VERI-Wild} & 156/2472 & 389/389 & 389/5985 \\
\textbf{Vehicle-1M} & 247/2579 & 611/611 & 611/6242 \\
\textbf{Total} & 403/5051 & 1000/1000 & 1000/12227 \\
\bottomrule
\end{tabular}
}
    \end{minipage}
\end{minipage}
  
  \end{figure}

\subsection{Implementation Details}
\noindent
\textbf{Training Stage.} \footnote{More details about training each stage and results are presented in the Supplementary Material.}For the proposed ReID network, ResNet-50 is adopted as the backbone, whose weights are initialized from the model pre-trained on the ImageNet. In the synthetic data training stage, the dimensions of FC layers are set to 3000. The weights of the domain transformation network are initialized by Kaiming normalization~\cite{he2015delving}. The whole network is trained in an end-to-end fashion based on the training sets of synthetic and real-world datasets for learning domain transformation, vehicle ReID and ID classification simultaneously. The input image is resized to 384 $\times$ 384. The training batch sizes at the synthetic data and the real-world data stages are 72 ($2M$) and 36 ($M$), respectively. The local patch size in the dark channel operation is $5\times5$. We apply the data augmentation in the training process including the random cropping and horizontal flipping techniques. The warm-up training strategy is adopted for 120 epochs. The Adam optimizer is adopted with a decay rate of 0.6. The initial learning rate is $1.09\times10^{-5}$, which increases to $10^{-4}$ after the $10^{th}$ epoch. At the training stage of the synthetic data, we adopt the synthetic dataset in \tabref{tab:syndataset} and randomly select one hazy image for each vehicle and put it into the probe set. The rest of images form the gallery set. For the training stage of real-world data, we apply 5051 clear images and hazy images without ID labels, respectively. The network is trained on an Nvidia Tesla V100 GPU for 3 days and we implement it using Pytorch.
\noindent \smallskip\\
\textbf{Inference Stage.} At the inference stage, the encoder ($\mathbf{E_{C}}$) and two decoders of the domain transformation network ($\mathbf{D_{C}}$ and $\mathbf{D_{H}}$) are not involved. The computational burden caused by them can be ignored. The Euclidean distance $D$ is computed through embedding features to evaluate the performance.

\begin{figure}[t!]
  \begin{minipage}[t]{0.5\textwidth}
          \captionof{table}{\textbf{Quantitative evaluation on the hazy vehicle ReID scenario.} The words with \textbf{boldface} indicate the best results, and those with \underline{underline} indicate the second-best results. The texts 'S' and 'R' indicate synthetic and real-world datasets.}

      \label{tab:quantitative}
    \centering
\scalebox{0.65}{\begin{tabular}{ccccccccc} 
\toprule
\multirow{2}{*}{\textbf{Method}} & \multicolumn{2}{c}{\textbf{mAP}} & \multicolumn{2}{c}{\textbf{CMC@1}} & \multicolumn{2}{c}{\textbf{CMC@5}} & \multicolumn{2}{c}{\textbf{CMC@10}} \\ 
\cline{2-9}
 & \textbf{S} & \textbf{R} & \textbf{S} & \textbf{R} & \textbf{S} & \textbf{R} & \textbf{S} & \textbf{R} \\ 
\hline
\textbf{VRCF} & 25.90 & 36.60 & 61.70 & 63.70 & 76.50 & 78.80 & 81.30 & 83.20 \\
\textbf{VRCF-dehaze} & 61.50 & 50.80 & 85.40 & 78.00 & 95.10 & 92.00 & 97.20 & 95.40 \\
\textbf{VRCF-haze} & 69.00 & 58.00 & 88.60 & 81.10 & 97.60 & 93.80 & 98.40 & 96.80  \\
\textbf{VRCF-all} & 73.00 & 64.40 & 90.90 & 85.50 & 97.30 & 95.70 & 98.60 & 97.80 \\
\textbf{VOC} & 59.70 & 57.40 & 86.10 & 82.80 & 94.30 & 94.00 & 95.60 & 96.60 \\
\textbf{VOC-dehaze} & 63.40 & 49.20 & 87.00 & 74.10 & 94.80 & 89.90 & 96.50 & 94.30 \\
\textbf{VOC-haze} & 67.10 & 59.90 & 88.70 & 83.50 & 95.10 & 94.00 & 96.50 & 97.20  \\
\textbf{VOC-all} & 84.20 & 78.70 & 93.60 & 91.00 & 97.60 & 96.30 & 98.30 & 98.30 \\
\textbf{DMT} & 73.90 & 71.70 & 93.40 & 93.20 & 97.20 & 97.40 & 97.90 & 98.50 \\
\textbf{DMT-dehaze} & 75.10 & 71.60 & 93.40 & 92.40 & 96.90 & 97.50 & 98.30 & 98.40 \\
\textbf{DMT-haze} & 77.30 & 73.40 & 94.00 & 93.40 & 97.60 & 97.60 & 98.60 & 98.80 \\
\textbf{DMT-all} & 82.50 & 80.90 & 98.30 & 96.10 & 98.20 & 98.20 & 98.80 & 99.00 \\
\textbf{VehicleX} & 63.64 & 61.56 & 86.50 & 83.20 & 95.00 & 95.20 & 97.40 & 97.90 \\
\textbf{VehicleX-dehaze} & 73.06 & 64.82 & 89.70 & 83.90 & 96.70 & 95.10 & 98.20 & 97.60 \\
\textbf{VehicleX-haze} & 77.86 & 69.01 & 91.20 & 84.80 & 97.10 & 96.10 & 98.70 & 98.10  \\
\textbf{VehicleX-all} & 80.75 & 76.39 & 93.10 & 89.90 & 97.60 & 96.90 & 98.60 & 98.40 \\
\textbf{TransReID} & 62.90 & 64.00 & 82.40 & 77.70 & 92.30 & 88.80 & 98.40 & 94.00 \\
\textbf{TransReID-dehaze} & 66.80 & 65.30 & 83.00 & 76.60 & 94.10 & 89.90 & 98.10 & 94.60 \\
\textbf{TransReID-haze}  & 73.90 & 72.10 & 84.80 & 82.60 & 95.20 & 90.70 & 98.70 & 95.60  \\
\textbf{TransReID-all} & 79.20 & 76.90 & 89.40 & 84.50 & 96.80 & 93.20 & 98.90 & 97.30 \\
\textbf{PVEN} & 72.83 & 75.36 & 63.73 & 66.48 & 84.39 & 86.53 & 89.65 & 91.20 \\
\textbf{PVEN-dehaze} & 81.70 & 78.13 & 73.29 & 69.47 & 92.50 & 89.16 & 96.04 & 93.43 \\
\textbf{PVEN-haze}  & 84.55 & 81.92 & 76.60 & 74.09 & 95.02 & 92.15 & 97.84 & 95.66  \\
\textbf{PVEN-all} & \underline{88.63} & \underline{84.08} & 83.55 & 78.31 & 98.45 & 95.40 & 99.20 & 97.76 \\
\textbf{HRCN} & 81.22 & 71.77 & 92.00 & 85.30 & 97.60 & 95.40 & 99.10 & 97.50 \\
\textbf{HRCN-dehaze} & 83.44 & 72.78 & 92.20 & 84.60 & 98.00 & 96.10 & 99.00 & 97.80 \\
\textbf{HRCN-haze} & 85.40 & 78.64 & 92.80 & 89.40 & \underline{98.50} & 96.70 & 99.10 & 98.40 \\
\textbf{HRCN-all} & 87.91 & 81.41 & 94.60 & 91.80 & 98.20 & 97.30 & \textbf{99.30} & 99.00 \\
\textbf{CAL} & 75.52 & 75.94 & 92.50 & 91.70 & 96.50 & 97.60 & 97.90 & 98.40 \\
\textbf{CAL-dehaze} & 83.21 & 77.49 & 94.80 & 94.00 & 98.30 & 98.00 & 98.90 & 98.80 \\
\textbf{CAL-haze} & 86.00 & 80.31 & 95.00 & 94.20 & 97.90 & \underline{98.30} & 98.90 & \underline{99.10} \\
\textbf{CAL-all} & 88.20 & 83.84 & \underline{96.30} & \textbf{96.00} & 98.40 & 98.20 & 98.90 & 99.00 \\
\hline
\textbf{Ours} & \textbf{88.66}
 & \textbf{84.12}
 & \textbf{96.70}
 & \underline{95.60}
 & \textbf{98.60}
 & \textbf{98.60}
 & \textbf{99.30}
 & \textbf{99.30}
 \\
\textbf{Ours-F} & \textbf{89.14}
 & \textbf{87.72}
 & \textbf{96.50}
 & \textbf{96.90}
 & \textbf{98.60}
 & \textbf{98.40}
 & \textbf{99.40}
 & \textbf{99.60}
 \\

\bottomrule
\end{tabular}
}
  \end{minipage}
  \hspace{0.5em}
\begin{minipage}[t]{0.7\textwidth}
  \begin{minipage}[t]{0.7\textwidth}
    \captionof{table}{\textbf{Ablation study for each module in the real-world test set.}}
        \centering
        \scalebox{0.7}{\begin{tabular}{ccccc} 
\toprule
\multirow{2}{*}{\textbf{Method}} & \multicolumn{4}{c}{\textbf{Metric}} \\ 
\cline{2-5}
 & \textbf{mAP} & \textbf{CMC@1} & \textbf{CMC@5} & \textbf{CMC@10} \\ 
\hline\hline
\textbf{Baseline-haze} & 76.17 & 93.40 & 97.50 & 98.50 \\
\textbf{Baseline-all} & 77.34 & 93.40 & 97.60 & 98.60 \\
\cdashline{1-5}
\textbf{Ours w/o} $\mathcal{L}_{CR}$\& $\mathcal{L}_{MIDC}$ & 76.02 & 93.50 & 97.20 & 98.60 \\
\textbf{Ours w/o} $\mathcal{L}_{MIDC}$ & 81.19 & 94.20 & 98.20 & 99.00 \\
\textbf{Ours w/o} $\mathcal{L}_{CR}$ & 82.27 & 95.30 & 98.00 & 99.20 \\
\cdashline{1-5}
\textbf{Ours w/o} $\mathcal{L}_{DC}$\&$\mathcal{L}_{TV}$ & 77.31 & 93.63 & 97.38 & 98.80 \\
\textbf{Ours w/o} $\mathcal{L}_{DC}$ & 81.20 & 94.30 & 98.30 & 98.85 \\
\textbf{Ours w/o} $\mathcal{L}_{TV}$ & 82.50 & 94.60 & 98.50 & 98.90 \\
\cdashline{1-5}
\textbf{Ours}& \textbf{84.12} & \textbf{95.60}  & \textbf{98.60}  & \textbf{99.30} \\
\bottomrule
\end{tabular}

}
    \label{tab:ablation}
    \end{minipage}
\\  
\hfill

 \begin{minipage}[t]{0.7\textwidth}
    \captionof{table}{\textbf{Comparison of performance for using different blocks as encoders $\mathbf{E_{C}}$ and $\mathbf{E_{H}}$ in the real-world test set.}}
        \centering
        \scalebox{0.8}{\begin{tabular}{ccccc} 
\toprule
\multirow{1}{*}{\textbf{}} & \multicolumn{1}{c}{\textbf{mAP}} & \multicolumn{1}{c}{\textbf{CMC@1}} & \multicolumn{1}{c}{\textbf{CMC@5}} & \multicolumn{1}{c}{\textbf{CMC@10}} \\ 
\hline\hline
\textbf{Conv\_2}  & \textbf{84.12} & \textbf{95.60}  & \textbf{98.60}  & \textbf{99.30} \\
\textbf{Conv\_3} & 83.84 & 94.90  & 98.20  & 98.90 \\
\textbf{Conv\_4}  & 82.56 & 94.40 & 98.20 & 99.00 \\
\textbf{Conv\_5} & 80.82 & 94.90 & 98.10 & 99.00 \\ 
\bottomrule
\end{tabular}

}
    \label{tab:convblock_compare}
    \end{minipage}

\leavevmode

\begin{minipage}[t]{0.7\textwidth}
    \captionof{table}{\textbf{Ablation study for using different training data stages in real-world test set.}}
        \centering
        \scalebox{0.75}{
        \begin{tabular}{ccccc} 
\toprule
\multirow{2}{*}{\textbf{Stage}} & \multicolumn{4}{c}{\textbf{Metric}} \\ 
\cline{2-5}
 & \textbf{mAP} & \textbf{CMC@1} & \textbf{CMC@5} & \textbf{CMC@10} \\ 
\hline\hline
\textbf{Syn} & 78.17 & 92.20 & 96.70 & 97.20  \\
\textbf{Syn+RC} & 80.03 & 94.10 & 98.20 & 98.90 \\
\textbf{Syn+RH} & 81.26 & 94.60 & 98.10 & 98.90  \\
\textbf{Ours}& \textbf{84.12} & \textbf{95.60}  & \textbf{98.60}  & \textbf{99.30} \\
\bottomrule
\end{tabular}

}
    \label{tab:multistage}
    \end{minipage}

    \end{minipage}
  \end{figure}

\begin{figure*}[t!]
\centering \includegraphics[width=1.0\textwidth]{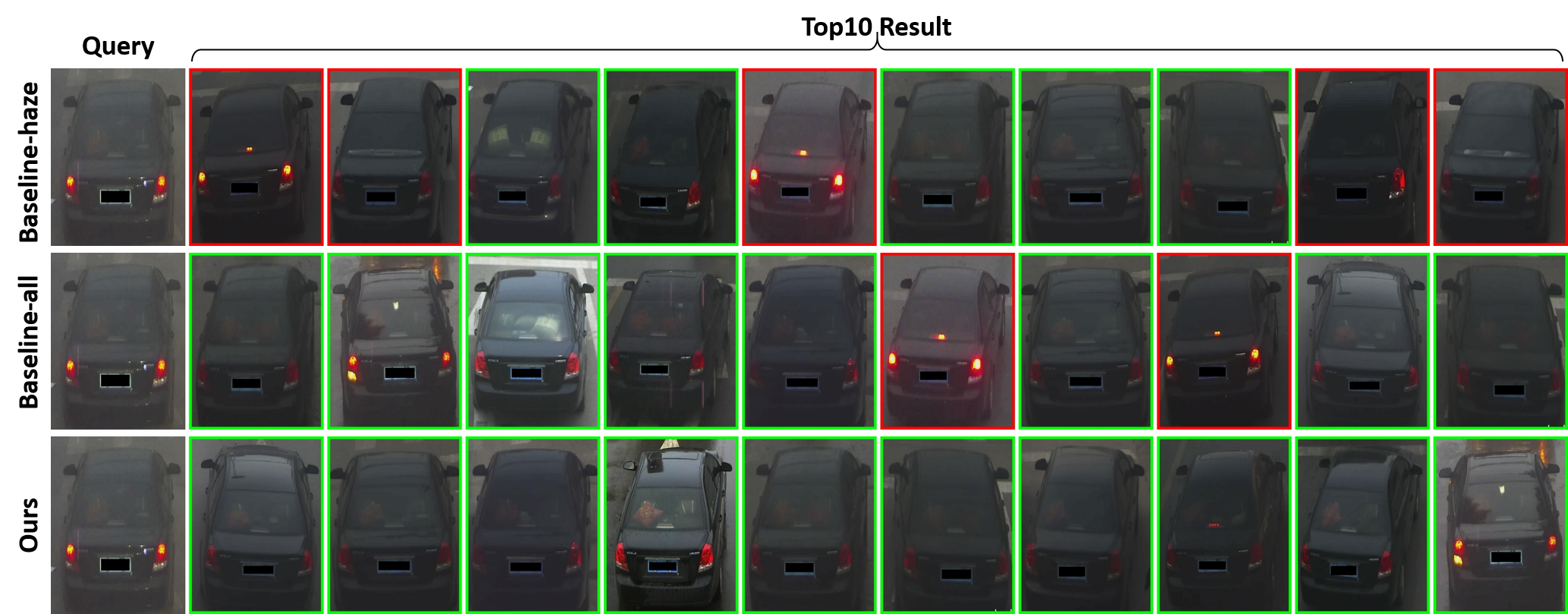}{}
\makeatother 
\caption{\textbf{Visual results of the ranking list on the real-world dataset.} The query images are in the first column and the retrieved top-10 ranking results are in the rest columns. We denote the correct retrieved images with a green border while the false instances are with a red border.}
\label{fig:rank_10}
\end{figure*}

\begin{figure}[t!]
\centering \includegraphics[width=0.8\textwidth]{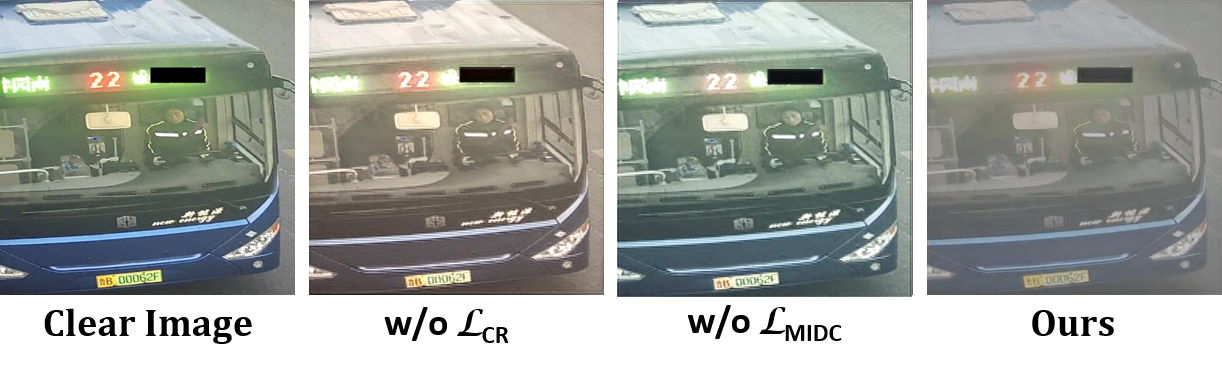}{}
\makeatother 
\caption{\textbf{Visual comparison of using the unsupervised losses $\mathcal{L}_{CR}$ and $\mathcal{L}_{MIDC}$ in the domain transformation network.} These loss functions can benefit the rendering process to generate more desirable results.}
\label{fig:visual_defog_compare}
\end{figure}

\subsection{Comparison with the Existing Methods}
To evaluate the performance of the proposed method, we compare the proposed algorithm with state-of-the-art ReID methods, the VRCF~\cite{gao2020vehicle}, the VOC ~\cite{zhu2020voc}, the DMT~\cite{he2020multi}, the CAL~\cite{rao2021counterfactual}, the VEHICLEX~\cite{yao2020simulating}, the TransReID~\cite{he2021transreid}, the PVEN~\cite{meng2020parsing}, and the HRCN~\cite{zhao2021heterogeneous}. For a fair and comprehensive comparison, these methods are retrained by the following training sets: (i) The ground truth clear images in the synthetic dataset; (ii) The hazy images from the training sets of both synthetic and real-world haze datasets (denoted with the '-haze'); (iii) The two-stage strategy (i.e., dehazing+ReID) which is denoted with '-dehaze'. Specifically, this strategy is the combination of the dehazing method for pre-processing and the ReID models trained by setting (i). For the dehazing method, we adopt one of the state-of-the-art dehazing methods called MPR-Net~\cite{zamir2021multi} which was retrained on hazy vehicle images. (iv) The same training images used in our method including synthetic haze, real-world clear and real-world haze datasets (denoted with '-all'). The aforementioned settings are all with complete ID labels.

The results are reported in~\tabref{tab:quantitative}. We can observe the following results. First, compared with other strategies, the proposed method can achieve the competitive performance on vehicle ReID in hazy weather on both synthetic and real-world datasets in terms of mAP and CMC. Second, existing methods trained on all data can obtain better performance compared to the methods trained on other training settings. Third, other methods may have limited performance in real-world scenarios, especially when they are only trained on synthetic images. Last, \textit{surprisingly, though our method is trained without ID labels in real-world data, it can outperform most supervised methods trained with complete ID labels}. 

We also adopt our method trained with ID labels in the real-world data training stage which is denoted with the suffix 'F'. Specifically, we introduce the triplet loss and the ID loss defined in \eqref{eq:idloss} and \eqref{eq:triploss} to train the real-world data stage. The result indicates that the performance can be improved if we use complete ID labels in the training stage. Our method can be also adopted in the fully supervised scenarios and obtain the decent performance.

\subsection{Ablation Studies}
\noindent
\textbf{Effectiveness of the Semi-supervised Strategy.} 
In this paper, we proposed the semi-supervised training technique to solve hazy vehicle ReID problem. It uses the domain transformation mechanism which enables us to train the real-world data without the ID labels. We present the effectiveness of this strategy in \tabref{tab:ablation}. We adopt the our ReID network as the baseline and train with two settings for comparison, that -is, the settings (ii) and (iv) reported in subsection 4.3 with complete ID labels. One can see that our method is against the first setting favorably. Moreover, even without the ID labels of real-world data, our method can outperform the baseline trained with complete ID labels since our methods integrate the domain transformation technique, which can improve the ReID network to learn better representation under the hazy scenes. We also show the visual results of the ranking list on the real-world dataset in \figref{fig:rank_10}. One can see that, the baseline retrieves wrong instances because the important features such as the window and light become ambiguous due to the degradation of haze, which may deteriorate the performance of ReID.

Moreover, in \tabref{tab:convblock_compare}, we show the results of assigning different convolution blocks as the encoders. One can see that adopting the first two convolution blocks as the encoders can obtain the best performance. However, the proposed domain transformation architecture can assist the network to learn accurate ReID in the haze scenario.
\noindent \smallskip\\
\textbf{Effectiveness of the Loss Functions.}
In \tabref{tab:ablation}, we verify the effectiveness of the adopted loss functions: the monotonously increasing dark channel loss $\mathcal{L}_{MICD}$ and the colinear relation constraint $\mathcal{L}_{CR}$. One can see that, with two loss functions, the performance of ReID can be improved in both mAP and CMC metrics since the DT-Net can benefit the encoders to learn more robust features with appropriate constraints which can further benefit the performance of ReID. Furthermore, using both the dark channel loss $\mathcal{L}_{DC}$ and the total variation loss $\mathcal{L}_{TV}$ can improve the performance of the network. \figref{fig:visual_defog_compare} presents that, with the proposed loss functions, the rendered results can be more realistic compared with other modules. The rendered results may have the color distortion problems without using $\mathcal{L}_{CR}$. 
\noindent \smallskip\\
\textbf{Effectiveness of Each Training Stage.}
We verify the effectiveness of using real clear data or real hazy data in the training process. We construct three settings for the comparison. Specifically, we adopt: (i) only the synthetic data stage (\textbf{Syn}), (ii) \textbf{Syn} with real clear data stage (\textbf{Syn+RC}), and (iii) \textbf{Syn} with real haze data stage (\textbf{Syn+RH}). The results are reported in \tabref{tab:multistage}. We can see that only adopting the synthetic data may cause limited performance in real-world scenarios due to the domain gap problem.

\section{Conclusion}
In this paper, to address the vehicle ReID problem under hazy scenarios, a semi-supervised training framework that integrates the domain transformation network and the ReID network is proposed. Moreover, to constrain the unsupervised training stage, several loss functions to bound the two networks are proposed. With these techniques, the proposed method can learn haze-invariant features for robust vehicle ReID. Experimental results show that, compared to existing methods trained on complete ID labels, the proposed methods can achieve decent performance even without using the ID labels in real-world data.

\section{Acknowledgement}
We thank to National Center for High-performance Computing (NCHC) for providing computational and storage resources. This research was supported by the Ministry of Science and Technology, Taiwan under Grants MOST 108-2221-E-002-072-MY3, MOST 108-2638-E-002-002-MY2, and MOST 111-2221-E-002-136-MY3.

%
%
\bibliographystyle{splncs04}
\bibliography{article}
\clearpage
\end{document}